# Robotics Applications in Neurology: A Review of Recent Advancements and Future Directions

Retnaningsih[†], Agus Budiyono[‡] and Rifky Ismail[‡‡], Dodik Tugasworo[£], Rivan Danuaji[††], Syahrul[$], Jerry Gunawan[¥]

[†]Department of Neurology, Kariadi Hospital, Indonesia.
[‡]Indonesia Center for Technology Empowerment (ICTE), Indonesia.
[‡‡] Center for Biomechanics Biomaterials Biomechatronics and Biosignal Processing, Diponegoro University, Indonesia.
[£]Faculty of Medicine, UNDIP, Semarang, Indonesia.
[††]Faculty of Medicine, UNS, Solo, Indonesia.
[$]Department of Neurology, Faculty of Medicine, Syiah Kuala University, Banda Aceh, Indonesia.
[¥]RS Permata Cirebon, Cirebon, Indonesia.

*Abstract:*

*Robotic technology has the potential to revolutionize the field of neurology by providing new methods for diagnosis, treatment, and rehabilitation of neurological disorders. In recent years, there has been an increasing interest in the development of robotics applications for neurology, driven by advances in sensing, actuation, and control systems. This review paper provides a comprehensive overview of the recent advancements in robotics technology for neurology, with a focus on three main areas: diagnosis, treatment, and rehabilitation. In the area of diagnosis, robotics has been used for developing new imaging techniques and tools for more accurate and non-invasive mapping of brain structures and functions. For treatment, robotics has been used for developing minimally invasive surgical procedures, including stereotactic and endoscopic approaches, as well as for the delivery of therapeutic agents to specific targets in the brain. In rehabilitation, robotics has been used for developing assistive devices and platforms for motor and cognitive training of patients with neurological disorders. The paper also discusses the challenges and limitations of current robotics technology for neurology, including the need for more reliable and precise sensing and actuation systems, the development of better control algorithms, and the ethical implications of robotic interventions in the human brain. Finally, the paper outlines future directions and opportunities for robotics applications in neurology, including the integration of robotics with other emerging technologies, such as neuroprosthetics, artificial intelligence, and virtual reality. Overall, this review highlights the potential of robotics technology to transform the field of neurology and improve the lives of patients with neurological disorders.*

Keywords: Robotics, Neurology, Diagnosis, Treatment, Rehabilitation.

**Introduction:**

Neurological disorders, such as stroke, Parkinson's disease, and spinal cord injury, affect millions of people worldwide and have a significant impact on patients' quality of life and healthcare systems. While conventional treatment approaches, such as medication and physical therapy, have been effective to some extent, they often have limited efficacy, high costs, and side effects. Moreover, the current healthcare system is facing challenges related to the increasing prevalence of neurological disorders and the aging population, leading to a growing demand for innovative and cost-effective solutions.

In recent years, robotics technology has emerged as a promising approach to address some of these challenges by providing new methods for diagnosis, treatment, and rehabilitation of neurological disorders. Robotics has the potential to offer more precise, reliable, and non-invasive interventions, as well as to enhance patient engagement, motivation, and feedback. Robotics can also enable more personalized and adaptive therapies based on patients' individual characteristics, leading to better outcomes and reduced healthcare costs.

The purpose of this review paper is to provide a comprehensive overview of the recent advancements in robotics technology for neurology and to discuss the potential and challenges of these applications. Specifically, the paper will focus on three main areas: diagnosis, treatment, and rehabilitation.

In the area of diagnosis, robotics technology has been used to develop new imaging techniques and tools for more accurate and non-invasive mapping of brain structures and functions. These technologies include robotic-assisted magnetic resonance imaging (MRI) and computed tomography (CT) scans, which allow for more precise and real-time imaging of the brain, as well as robotic devices for biopsy and tissue sampling, which enable more targeted and minimally invasive procedures.

In the area of treatment, robotics technology has been used for developing minimally invasive surgical procedures, including stereotactic and endoscopic approaches, as well as for the delivery of therapeutic agents to specific targets in the brain. Robotics can enable more precise and targeted interventions, leading to reduced surgical time, blood loss, and risk of complications. Robotics technology has also been used for developing neuroprosthetics, such as brain-computer interfaces and deep brain stimulation devices, which can restore or enhance neural function and communication.

In the area of rehabilitation, robotics technology has been used for developing assistive devices and platforms for motor and cognitive training of patients with neurological disorders. These devices include exoskeletons, end-effectors, and hybrid devices, which can provide personalized and intensive training modalities, as well as feedback and monitoring of patients' progress. Robotics technology can also enable the integration of virtual reality and gaming elements, which can enhance patient engagement and motivation, leading to better adherence and outcomes.

While robotics technology has shown great potential for neurology, there are still several challenges and limitations that need to be addressed. These include the need for more reliable and precise sensing and actuation systems, the development of better control algorithms, and the ethical implications of robotic interventions in the human brain. Moreover, there is a need for more rigorous clinical trials to establish the efficacy and safety of these technologies, as well as for addressing the concerns of patients, clinicians, and policymakers regarding the adoption and implementation of these technologies in clinical practice.

Robotics technology has the potential to transform the field of neurology by providing new methods for diagnosis, treatment, and rehabilitation of neurological disorders. The recent advancements in robotics technology have shown great promise for improving patient outcomes and reducing healthcare costs. However, there are still several challenges and limitations that need to be addressed, and more research is needed to establish the efficacy, safety, and feasibility of these technologies in clinical practice. This review paper aims to provide a comprehensive overview of the recent developments in robotics technology for neurology and to highlight the potential and challenges of these applications for future research and development.

**Literature review**

The use of robotic technology in neurological disorders has the potential to revolutionize diagnosis, treatment, and rehabilitation. Robotic devices have been successfully used for therapeutic and diagnostic purposes for stroke recovery, such as the robot-assisted therapy of the arm after stroke in a multicentre, parallel-group randomized trial [8]. Researchers found that robotic devices can effectively aid in stroke rehabilitation by providing patients with targeted, intensive, and repetitive training. These devices can provide real-time feedback, allowing for immediate correction of movement errors, which can promote more efficient and effective motor learning. By providing targeted and intensive training, they can promote more efficient and effective motor learning, while providing objective measures of performance and function. These devices can also be customized to suit the needs of individual patients and can be programmed to provide specific types of training or therapy.

Robot-assisted stereotactic biopsy has been shown to be effective in pediatric brainstem and thalamic lesions [7], as well as for frameless robot-assisted stereotactic biopsies for lesions of the brainstem [1]. Furthermore, robot-assisted procedures have been shown to be successful in pediatric neurosurgery [11]. Robot-assisted procedures have shown promise in pediatric neurosurgery due to their precision, accuracy, and ability to minimize damage to surrounding tissues. Robot-assisted procedures have shown potential in pediatric neurosurgery as a way to improve surgical precision and accuracy while minimizing damage to surrounding tissues. However, more studies are needed to fully assess the effectiveness and safety of this approach in this patient population.

Robotic devices have also been explored as a potential next-generation technology for clinical assessment of neurological disorders and upper-limb therapy [6]. Additionally, robot-assisted diagnosis of developmental coordination disorder has been investigated, although further research is needed in this area [3]. Furthermore, assistive technology and robotic control using motor cortex ensemble-based neural interface systems have shown promise in individuals with tetraplegia [10].

Robotic technology has also been used in the field of intensive care, where robotic telepresence has been shown to facilitate rapid physician response to unstable patients and decrease costs in neurointensive care [9]. Additionally, a study on neurological and robot-controlled induction of an apparition has been conducted, which explored the intersection between robotics and neuroscience [2].

Efficacy of rehabilitation robotics for walking training in neurological disorders has been reviewed, indicating positive outcomes [5]. The potential of robots in rehabilitation has been recognized, as they have been shown to be useful as therapeutic and diagnostic tools for stroke recovery, as well as for walking training in neurological disorders [1, 5, 8].

Overall, the use of robotic technology in neurological disorders has shown promise in various aspects of diagnosis, treatment, and rehabilitation. Robot-assisted procedures have been successful in pediatric neurosurgery, and robot-assisted stereotactic biopsy has been effective for lesions of the brainstem and thalamic lesions. Robotic devices have also been explored as a potential next-generation technology for clinical assessment of neurological disorders and upper-limb therapy, as well as for robot-assisted diagnosis of developmental coordination disorder. Additionally, assistive technology and robotic control using motor cortex ensemble-based neural interface systems have shown promise in individuals with tetraplegia. In the field of intensive care, robotic telepresence has been shown to facilitate rapid physician response to unstable patients and decrease costs. Further research is needed to fully explore the potential of robotic technology in the field of neurological disorders.

**Frontiers in Robotics Applications in Neurology**

**Robot-assisted diagnosis:** Robotics can be used in the diagnosis of neurological disorders, such as brain tumors or developmental coordination disorders, by providing high-resolution images or by measuring certain indicators of the disease. Robots can help clinicians to diagnose various diseases and disorders more accurately, quickly, and non-invasively than traditional methods. There are several advantages to using robot-assisted diagnosis, such as reducing human error, improving the efficiency of the diagnostic process, and providing more objective and consistent results.

One example of robot-assisted diagnosis is in the field of radiology, where robots can be used to analyze medical images, such as X-rays, MRIs, and CT scans. These robots can be programmed to detect abnormalities in the images, such as tumors, fractures, or other signs of disease. Using robots to analyze medical images can help to reduce the time required to make a diagnosis, while also improving the accuracy of the diagnosis.

Another example of robot-assisted diagnosis is in the field of pathology. Robots can be used to analyze tissue samples obtained during biopsy procedures. The robots can analyze the tissue samples for abnormalities, such as cancerous cells, and can provide a more accurate and objective diagnosis. This can help clinicians to develop a more effective treatment plan for the patient.

Robot-assisted diagnosis is also being explored for use in the diagnosis of neurological disorders, such as Alzheimer's disease, Parkinson's disease, and multiple sclerosis. In these cases, robots can be used to analyze brain scans and other diagnostic data to identify patterns and markers that are associated with these diseases. This can help clinicians to make an early and accurate diagnosis, which is critical for developing effective treatment plans.

One of the most promising areas of robot-assisted diagnosis is the use of artificial intelligence (AI) and machine learning algorithms to analyze medical data. These algorithms can be trained to recognize patterns and markers that are associated with specific diseases, which can help to improve the accuracy and speed of the diagnostic process. For example, machine learning algorithms have been used to analyze medical images to identify signs of breast cancer, and have been shown to be more accurate than human radiologists.

In conclusion, robot-assisted diagnosis has the potential to revolutionize medical diagnosis and improve patient outcomes. By leveraging the power of robots, AI, and machine learning algorithms, clinicians can make more accurate and efficient diagnoses, leading to better treatment plans and improved patient outcomes.

**Robot-assisted surgery**: Robot-assisted surgery refers to a surgical technique that uses robotic technology to aid the surgeon during the procedure. This technology typically involves a robotic arm that is controlled by the surgeon through a computer console. The robot's arms are equipped with tiny surgical instruments that can perform precise movements that are difficult to accomplish with human hands. This type of surgery is also sometimes referred to as computer-assisted surgery, robot-assisted laparoscopy, or robot-assisted minimally invasive surgery.

Robot-assisted surgery has several advantages over traditional surgery. First, it allows for greater precision and accuracy, which can result in less trauma to the patient's body. This is particularly important in delicate surgeries, such as neurosurgery, where even a small mistake can have serious consequences. Second, it allows for smaller incisions, which can lead to less scarring and a faster recovery time for the patient. Finally, robot-assisted surgery can be performed with greater efficiency, reducing the amount of time the patient spends in the operating room.

Some examples of robot-assisted surgery in neurological applications include:

- Brain surgery: Robot-assisted surgery can be used to remove brain tumors, perform biopsies, and treat other neurological conditions. The robot's precision and accuracy make it particularly useful in situations where the surgeon needs to avoid damaging healthy brain tissue.
- Spinal surgery: Robot-assisted surgery can be used to perform spinal fusions, decompressions, and other procedures. The robot's precision is particularly useful in spinal surgery, where even a small error can result in permanent nerve damage.
- Deep brain stimulation (DBS): DBS is a procedure used to treat Parkinson's disease, essential tremor, and other movement disorders. During the procedure, a small electrode is implanted in the patient's brain, and electrical signals are sent to the brain to control movement. Robot-assisted surgery can be used to precisely place the electrode, reducing the risk of complications.

Robot-assisted surgery has shown promising results in the field of neurological surgery. While the technology is still relatively new, it has the potential to improve patient outcomes and reduce the risks associated with traditional surgery.

**Robot-assisted rehabilitation:** Robot-assisted rehabilitation refers to the use of robotic devices to assist individuals in recovering from neurological or physical impairments caused by injuries, illnesses, or other conditions. The goal of robot-assisted rehabilitation is to help patients improve their motor function, regain mobility, and ultimately achieve greater independence in their daily lives.

Robot-assisted rehabilitation has become increasingly common in recent years, particularly for individuals recovering from strokes or spinal cord injuries. The devices used in robot-assisted rehabilitation are typically designed to provide assistance and support to patients as they perform exercises and movements aimed at improving their strength, coordination, and range of motion.

One example of a robot-assisted rehabilitation device is the Lokomat, a robotic exoskeleton designed to help individuals with lower limb paralysis learn to walk again. The Lokomat provides support to the patient's legs while also guiding their movements, allowing them to practice walking without fear of falling. Other examples of robot-assisted rehabilitation devices include robotic arm and hand exoskeletons that can help individuals recovering from upper limb injuries or disorders, and virtual reality systems that can provide patients with engaging and motivating exercise environments.

Robot-assisted rehabilitation has several advantages over traditional physical therapy. First, the use of robotics can provide more consistent, precise, and repeatable movements and exercises, which can be important for promoting optimal recovery. Second, robot-assisted rehabilitation can provide patients with immediate feedback on their performance, allowing them to adjust their movements and make progress more quickly. Third, robotic devices can provide patients with a greater degree of safety and support during exercises, reducing the risk of falls or other injuries.

Robot-assisted rehabilitation has the potential to significantly improve outcomes for patients recovering from neurological or physical impairments. By providing more consistent, precise, and engaging exercise environments, robotic devices can help patients achieve greater gains in motor function and ultimately achieve greater independence in their daily lives.

**Neuroprosthetics**: Neuroprosthetics is a field that focuses on the development of devices that can replace or restore damaged sensory, motor, or cognitive function in the nervous system. Neuroprosthetic devices work by interfacing with the nervous system at various levels, from the peripheral nerves to the brain, to restore or augment neural function. These devices can take different forms, including implanted electrodes, nerve cuffs, and brain-computer interfaces (BCIs).

One of the most widely used neuroprosthetic devices is the cochlear implant, which is a surgically implanted electronic device that provides auditory input to the brain of people with severe hearing loss. Cochlear implants work by converting sound waves into electrical signals that stimulate the auditory nerve, bypassing the damaged hair cells in the cochlea. Another example of a neuroprosthetic device is the deep brain stimulation (DBS) system, which is used to treat movement disorders such as Parkinson's disease, tremors, and dystonia. DBS involves implanting electrodes in specific areas of the brain and delivering electrical impulses to modulate neural activity and improve motor function.

Brain-computer interfaces (BCIs) are another form of neuroprosthetics that allow direct communication between the brain and a computer or other external devices. BCIs work by recording the electrical activity of the brain using implanted or non-invasive electrodes and translating it into commands that can control a robotic arm, a wheelchair, or a computer cursor, among other applications. BCIs can be used to restore lost motor function in people with spinal cord injuries, ALS, or stroke, as well as to enhance cognitive function in individuals with neurological disorders such as epilepsy or ADHD.

Other examples of neuroprosthetic devices include retinal implants, which are used to restore vision in people with retinitis pigmentosa or age-related macular degeneration, and prosthetic limbs, which can restore mobility and functionality in amputees. In recent years, there has been a growing interest in the development of soft and flexible neuroprosthetic devices that can conform to the shape and movements of the body and provide a more natural interface with the nervous system.

Despite the significant progress made in the field of neuroprosthetics, there are still many challenges to overcome, such as improving the longevity and stability of implanted devices, minimizing the risk of infection and tissue damage, and optimizing the integration of artificial and biological components. Nevertheless, the potential benefits of neuroprosthetics for people with neurological disorders or injuries are immense, and ongoing research in this area is likely to lead to further advances in the years to come.

**Robot-assisted telemedicine**: Robot-assisted telemedicine refers to the use of robotic technologies to deliver medical care remotely to patients. This approach has gained significant attention in recent years due to its potential to provide healthcare services to remote and underserved areas, increase access to specialists, and reduce healthcare costs.

Robot-assisted telemedicine systems can be divided into two categories: stationary and mobile systems. Stationary systems are typically located in hospitals or clinics and are used to remotely monitor patients in real-time, as well as to provide consultation and education to healthcare providers. Mobile systems, on the other hand, are designed to be transported and used in various locations, including patients' homes, remote clinics, and disaster sites.

One of the most common applications of robot-assisted telemedicine is telepresence, which allows healthcare providers to remotely assess and interact with patients. For instance, Intensive Care Unit

(ICU) robotic telepresence systems have been developed to enable remote monitoring and management of critically ill patients, particularly those in remote and underserved areas. Such systems allow physicians to remotely access patient information, provide timely interventions, and even perform remote consultations with specialists.

Robot-assisted telemedicine can also be used for remote consultations and diagnosis. For example, in pediatric neurosurgery, robot-assisted procedures have been shown to be successful in reducing the risk of surgical complications and increasing the accuracy of diagnosis. Similarly, robot-assisted diagnosis can be used in the early detection of diseases, such as cancer, by analyzing medical images and providing real-time feedback to healthcare providers.

Another application of robot-assisted telemedicine is in remote rehabilitation. For instance, robotic exoskeletons have been developed to assist patients in regaining mobility following a neurological injury, such as a stroke. These devices use sensors to detect the patient's movements and provide the necessary support and resistance to aid in rehabilitation. Moreover, these systems can be remotely monitored by healthcare providers, who can adjust the rehabilitation program as needed based on the patient's progress.

Robot-assisted telemedicine is a rapidly growing field that has the potential to revolutionize healthcare delivery. With advances in robotics technology, remote medical care can be made more accessible, efficient, and effective, especially for patients in remote or underserved areas. Robot-assisted telemedicine has already shown promising results in a variety of applications, including telepresence, remote consultations and diagnosis, and remote rehabilitation.

**Brain-computer interfaces (BCIs):** Brain-computer interfaces (BCIs) are devices that enable direct communication between the brain and a computer, allowing individuals to control a device or communicate with the outside world through their thoughts. BCIs work by translating signals from the brain into computer commands using electrodes that are placed on or in the brain.

There are several types of BCIs, including invasive, partially invasive, and non-invasive BCIs. Invasive BCIs involve surgically implanting electrodes directly into the brain tissue, while partially invasive BCIs use electrodes implanted in the skull or on the surface of the brain. Non-invasive BCIs, such as electroencephalography (EEG) or functional magnetic resonance imaging (fMRI) BCIs, measure brain activity through the scalp or indirectly through blood flow.

One of the most common uses of BCIs is to restore communication and movement to individuals with severe disabilities, such as amyotrophic lateral sclerosis (ALS), brainstem stroke, and spinal cord injury. For example, a person with ALS who has lost the ability to move their limbs and speak can use a BCI to control a computer cursor, select letters or words, and generate speech using a text-to-speech synthesizer. This can significantly improve their quality of life and enable them to communicate with others.

BCIs have also been used to control robotic prosthetic limbs, allowing individuals with upper limb amputations to regain the ability to perform everyday tasks such as grasping and manipulating objects. The prosthetic limb is controlled by the individual's thoughts, which are detected by the BCI and used to activate motors in the prosthetic limb.

Another potential application of BCIs is in the field of neuromodulation, which involves the use of electrical or magnetic stimulation to modulate brain activity. BCIs can be used to detect abnormal brain activity, such as seizures or tremors, and provide feedback or stimulation to restore normal brain function.

BCIs are also being explored as a tool for cognitive enhancement, such as improving attention or memory. For example, a BCI system could be used to monitor brain activity and provide feedback to help individuals improve their focus or memory recall.

In addition to clinical applications, BCIs are also being used in research to better understand the brain and its functions. Researchers can use BCIs to record and analyze brain activity during various tasks, providing insights into how the brain processes information and controls behavior.

While BCIs have shown great potential in restoring function to individuals with disabilities and improving our understanding of the brain, there are still significant challenges to overcome, such as improving the accuracy and reliability of the technology, and ensuring the long-term safety of implanted devices. Nonetheless, BCIs represent a rapidly advancing field with the potential to transform the lives of individuals with neurological disorders or injuries.

Discussion

**Overview of Current Progress**

Recent advancements in robotics have led to promising applications in the field of neurology, including diagnosis, treatment, and rehabilitation of neurological disorders. Robot-assisted diagnosis has shown great potential in improving the accuracy and speed of diagnosing neurological disorders. Robotic systems can assist physicians in detecting small lesions in the brainstem or thalamus, which are difficult to access through traditional diagnostic methods, by performing stereotactic biopsies. Additionally, robot-assisted diagnosis has shown potential in diagnosing developmental coordination disorder, which could help to improve early diagnosis and intervention for affected individuals. Robotic devices have been utilized for diagnostic purposes, such as robot-assisted biopsies of brainstem and thalamic lesions, as well as for telemedicine, facilitating remote consultations and examinations by physicians.

Robot-assisted surgery has revolutionized neurosurgery by providing surgeons with greater accuracy and precision during delicate procedures. Robotic systems can be used to perform complex procedures in difficult-to-reach areas of the brain with minimal damage to surrounding tissue. This technology has been shown to be effective in procedures such as stereotactic biopsies, deep brain stimulation, and tumor resection. Robotic-assisted surgery has also been successful in pediatric neurosurgery, allowing for precise and minimally invasive procedures. In terms of treatment, robotic devices have been shown to be effective tools for stroke recovery, providing task-specific and individualized therapy.

Additionally, neuroprosthetics and brain-computer interfaces have emerged as promising options for restoring function and improving quality of life in patients with neurological disabilities. Neuroprosthetics have the potential to restore lost sensory or motor function in individuals with neurological injuries or diseases. Advances in technology have led to the development of prosthetic limbs that can be controlled by the user's thoughts, providing greater freedom and independence. In addition, neuroprosthetics can be used to stimulate neural circuits and improve cognitive function. Meanwhile, brain-computer interfaces have the potential to revolutionize the field of neurology by providing a direct link between the brain and external devices. This technology has shown potential in improving communication and motor function in individuals with severe disabilities. In addition, brain-computer interfaces can be used to treat neurological disorders such as Parkinson's disease, by stimulating specific brain regions to reduce symptoms. While these advancements are promising, there is still a need for further research and development to optimize the use of robotic technology in neurology. Future directions may include the development of more advanced and versatile robotic devices, as well as the integration of artificial intelligence and machine learning algorithms to enhance diagnostics and treatment. Overall, robotics has the potential to revolutionize the field of neurology, offering new avenues for improving patient outcomes and quality of life.

**Development of robotic application in Indonesia**

The development of robots for medical devices applications in Indonesia is growing quite rapidly. Several universities are working with hospitals to develop health technology robots. These robots were developed to assist patients and doctors in carrying out their duties. One of the universities discussed in this article is Diponegoro University, which has a center of excellence in the field of health technology. The centre of excellence, known as the centre for biomechanics, biomaterials, biomechatronics and biosignal processing (CBIOM3S) in Diponegoro University, develops robots for diagnosis, therapy, neuroprosthetics and brain computer interfaces.

In the field of robots for diagnostics, CoE CBIOM3S UNDIP has developed a Parkinson's diagnostic tool based on patient voice signal samples and muscle contraction signals (Putri, et al., 2018). The Parkinson's detection method in this study uses the pattern recognition method, the first step begins with the acquisition of voice and EMG data. The second step is feature extraction using five features for sound and EMG signals. The final stage is classification using the Adaptive Neuro-Fuzzy Inference System (ANFIS) and Artificial Neural Network (ANN) methods. Several patients from dr. Kariadi hospital had been diagnosed using this method. The result consists of: (i) the two-class classification (normal and definite Parkonson's) and (ii) the four-class classification (normal, probable, possible, definite Parkonson's). The first one has a higher accuracy than in both ANN and ANFIS. Based on the classification results of the four classes on sound and EMG signals using ANN and ANFIS, the probable class has the lowest accuracy of all classes. These methods had been embedded on a portable device which easily carried out by a doctor.

In the field of robotic therapy, CoE CBIOM3S UNDIP developed medical devices in collaboration with several hospitals, one of which was Dr. kariadi. Some of these tools are: Robot-Assisted Exoskeleton Therapy for Improving Muscle Strength on lower elbow and upper elbow muscles (Alfaina, et al., 2021 and Pangesti, et al., 2021), soft elbow exoskeleton for upper limb movement assistance (Ismail, 2019), soft exoskeleton glove using a motor-tendon actuator for finger therapy and hand movement assistance (Setiawan, et al., 2020 and Setiawan, et al., 2021a) and Extra robotic thumb and exoskeleton robotic fingers for patients with hand function disabilities (Ismail, et al., 2017). This therapy robot is still being continued to help patients in the elbows and fingers.

Neuroprosthetic is the most advanced type of robot made by CoE CBIOM3S Diponegoro University. Its superior product is a bionic hand for transradial amputation patients (Ismail, et al., 2018 and Ariyanto, et al., 2019a). This bionic hand product has been used by several amputee patient respondents to test its reliability. The process of developing this tool is accompanied by measurements of the function of hand movements which can see how bionic hands can help carry out activities of daily living (Susanto, et al., 2018) also accompanied by reliable finger movement performance tests so that bionic hands can last a long time (Pambudi, et al., 2019).

In the field of computer brain interfaces, one of the latest developments is the mechanism for controlling Flexion and Extension for Prosthetic Hand Controlled by Single-Channel EEG (Setiawan, et al., 2021b). The Diponegoro University CBIOM3S CoE developed EEG as part of a non-invasive BCI, in which electrodes must be placed outside the skull or on the scalp. Signal processing for EEG is quite complex and requires advanced calculation methods. The bionic hand created by CoE CBIOM3S is controlled with brain waves to incorporate movement feedback with potentiometers to the finger joints.

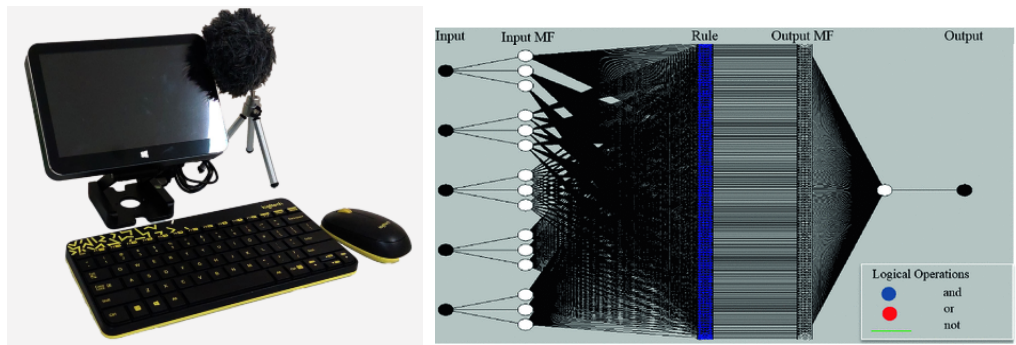

Fig. 1 Parkinson's early detection device and methods

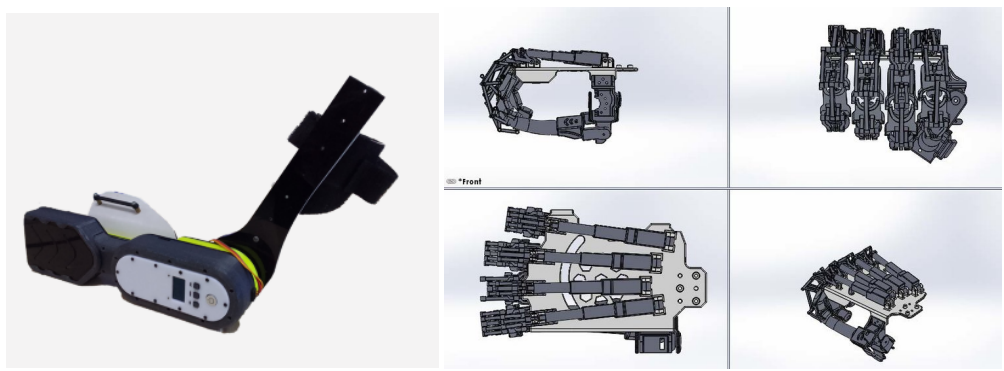

Fig. 2 Elbow exoskeletons and the finger exoskeleton to assist the patients upper limb movements.

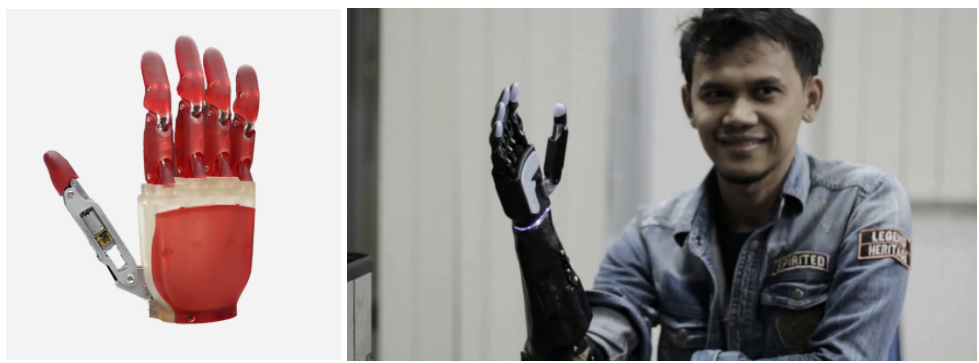

Fig. 3 Diponegoro bionic hand and the application on the trans-radial patients

**Conclusion**

The field of robotics has made significant advancements in neurology, offering promising applications in the diagnosis, treatment, and rehabilitation of neurological disorders. Robot-assisted diagnosis has shown potential in improving the accuracy and speed of diagnosing neurological conditions, particularly in detecting small lesions in challenging areas like the brainstem or thalamus. Additionally, robot-assisted surgery has revolutionized neurosurgery by providing surgeons with greater precision and minimizing damage to surrounding tissues. This technology has been successfully employed in complex procedures such as stereotactic biopsies, deep brain stimulation, and tumor resection.

Furthermore, robotic devices have proven effective in stroke recovery, providing personalized and task-specific therapy for improved rehabilitation outcomes.

The development of neuroprosthetics and brain-computer interfaces has also offered new possibilities for restoring function and enhancing the quality of life for individuals with neurological disabilities. Neuroprosthetics can restore sensory or motor function through the use of prosthetic limbs controlled by the user's thoughts. This technology has advanced to provide greater freedom and independence to individuals with limb loss. Brain-computer interfaces establish a direct link between the brain and external devices, enabling improved communication and motor function for those with severe disabilities. These interfaces can also be used to treat specific neurological disorders, such as Parkinson's disease, by stimulating targeted brain regions to alleviate symptoms.

However, further research and development are still necessary to optimize the use of robotic technology in neurology. Future directions may include the advancement of more sophisticated and versatile robotic devices, as well as the integration of artificial intelligence and machine learning algorithms to enhance diagnostics and treatment. The development of such technologies should prioritize usability, affordability, and accessibility to ensure widespread implementation and benefit to patients.

Moreover, in the context of Indonesia, significant progress has been made in the development of robotics for medical applications. Collaborations between universities and hospitals, such as the Centre for Biomechanics, Biomaterials, Biomechatronics, and Biosignal Processing (CBIOM3S) at Diponegoro University, have led to the creation of innovative robots for diagnosis, therapy, neuroprosthetics, and brain-computer interfaces. These advancements have included the development of diagnostic tools for Parkinson's disease based on voice and muscle contraction signals, as well as various robotic devices for therapy, including exoskeletons for elbow and finger movement assistance, soft exoskeleton gloves for finger therapy, and a bionic hand for transradial amputation patients. These developments demonstrate the potential for robotics to enhance healthcare in Indonesia and improve patient outcomes.

In conclusion, the integration of robotics into neurology has shown remarkable progress and holds great promise for the future. The advancements in diagnosis, surgery, rehabilitation, neuroprosthetics, and brain-computer interfaces have the potential to revolutionize the field, offering new avenues for improved patient care and quality of life. Continued research and collaboration are essential to further refine and expand these technologies, making them more accessible and beneficial to a broader range of patients. With the continuous development and application of robotics in neurology, we can look forward to a future where patients with neurological disorders receive more accurate diagnoses, precise treatments, and effective rehabilitation, ultimately leading to improved outcomes and a higher quality of life.